\documentclass[leqno, 12pt]{article}
\usepackage{arydshln}

\usepackage{longtable}

\usepackage{indentfirst}
\usepackage{amsfonts}
\usepackage{ntheorem}
\usepackage{amsmath}
\usepackage{epigraph}
\usepackage{bbm}
\usepackage{sectsty}
\usepackage[norule,bottom]{footmisc}
\usepackage[justification=centering,labelfont={bf}]{caption}
\usepackage{varioref}
\usepackage[letterpaper, margin=1in]{geometry}
\usepackage{amssymb}
\usepackage{graphicx}
\usepackage{float}
\usepackage{amsmath}
\usepackage{multirow}
\usepackage{setspace}
\usepackage{caption}
\usepackage{lscape}
\usepackage{subcaption}
\usepackage{xcolor}
\usepackage{xspace}
\usepackage{tabularx}
\usepackage{tabu}
\usepackage{booktabs}
\usepackage{afterpage}
\usepackage[utf8]{inputenc}
\usepackage{enumitem}
\usepackage[bottom]{footmisc}
\usepackage{pdflscape, lscape}
\usepackage[colorlinks=true, linkcolor=blue, citecolor = blue, bookmarks=false, urlcolor=blue]{hyperref}
\usepackage{comment}
\usepackage{lipsum,atbegshi}
\usepackage{tikz}  
\usetikzlibrary{arrows.meta} 
\usepackage[backend=biber,style=trad-abbrv]{biblatex}
\usepackage{hyperref}

\makeatletter
\renewbibmacro*{cite:plabelyear+extradate}{%
  \iffieldundef{labelyear}{}
    {\clearfield{labelmonth}
     \clearfield{labelday}
     \clearfield{labelendmonth}
     \clearfield{labelendday}
     \iffieldsequal{labelyear}{labelendyear}
       {\clearfield{labelendyear}}
       {}%
     \iffieldundef{origyear}
       {}
       {\printorigdate%
        \setunit*{\addslash}}%
     \iffieldundef{related}
       {}
       {\iffieldequalstr{relatedtype}{reprintfrom}
         {\entrydata*{\thefield{related}}{\printlabeldateextra}%
          \setunit*{\addslash}}
         {}}%
     \printlabeldateextra}}

\renewbibmacro*{cite}{%
  \iffieldequals{fullhash}{\cbx@lasthash}
   {\setunit{\compcitedelim}%
    \printtext[bibhyperref]{%
      \usebibmacro{cite:plabelyear+extradate}}}%
   {\printtext[bibhyperref]{%
      \ifnameundef{labelname}
       {\usebibmacro{cite:noname}%
         \setunit{\printdelim{nameyeardelim}}%
         \usebibmacro{cite:plabelyear+extradate}%
         \savefield{fullhash}{\cbx@lasthash}}
       {\ifnameundef{shortauthor}
         {\printnames{labelname}}%
         {\cbx@apa@ifnamesaved
           {\printnames{shortauthor}}
           {\ifnameundef{groupauthor}
             {\printnames[labelname]{author}}
             {\printnames[labelname]{groupauthor}}%
            \addspace\printnames[sabrackets]{shortauthor}}}%
         \setunit{\printdelim{nameyeardelim}}%
        \usebibmacro{cite:plabelyear+extradate}%
        \savefield{fullhash}{\cbx@lasthash}}}}%
   \setunit{\multicitedelim}}

\renewbibmacro*{textcite}{%
  \iffieldequals{fullhash}{\cbx@lasthash}
    {\setunit{\compcitedelim}%
     \printtext[bibhyperref]{%
       \usebibmacro{cite:plabelyear+extradate}}}
    {%
    \ifbool{cbx:parens}
      {\bibcloseparen\global\boolfalse{cbx:parens}}
      {}%
      \setunit{\compcitedelim}%
      \ifnameundef{labelname}
       {\iffieldundef{shorthand}%
         {\printtext[bibhyperref]{%
            \usebibmacro{cite:noname}}%
          \setunit{\ifbool{cbx:np}%
                   {\printdelim{nameyeardelim}}%
                   {\global\booltrue{cbx:parens}\addspace\bibopenparen}}%
          \printtext[bibhyperref]{%
            \usebibmacro{cite:plabelyear+extradate}}}
         {\printtext[bibhyperref]{%
            \usebibmacro{cite:shorthand}}}}
       {\printtext[bibhyperref]{%
          \ifnameundef{shortauthor}%
           {\printnames{labelname}}
           {\cbx@apa@ifnamesaved
             {\printnames{shortauthor}}
             {\ifnameundef{groupauthor}
               {\printnames[labelname]{author}}
               {\printnames[labelname]{groupauthor}}}}}%
        \setunit{\ifbool{cbx:np}
                  {\printdelim{nameyeardelim}}
                  {\global\booltrue{cbx:parens}\addspace\bibopenparen}}%
        \printtext[bibhyperref]{%
          \ifnameundef{shortauthor}
           {}
           {\cbx@apa@ifnamesaved
             {}
             {\printnames{shortauthor}\setunit{\printdelim{nameyeardelim}}}}%
          \usebibmacro{cite:plabelyear+extradate}}%
        \savefield{fullhash}{\cbx@lasthash}}}}
\makeatother

\AtBeginDocument{}
\newcolumntype{A}{>{\arraybackslash}m{5cm}}
\newcolumntype{B}{>{\centering\arraybackslash}m{2.1cm}}
\newcolumntype{C}{>{\centering\arraybackslash}m{0.5cm}}
\newcolumntype{D}{>{\centering\arraybackslash}m{3.9cm}}
\newcolumntype{E}{>{\centering\arraybackslash}m{3.2cm}}
\newcolumntype{F}{>{\centering\arraybackslash}m{2.31cm}}
\newcolumntype{L}{>{\arraybackslash}m{7.4cm}}
\theoremindent\parindent
\makeatletter
\renewtheoremstyle{plain}{\item[\hskip\labelsep\hskip-\parindent \theorem@headerfont ##1\ ##2\theorem@separator]}
{\item[\hskip\labelsep\hskip-\parindent \theorem@headerfont ##1\ ##2\ (##3)\theorem@separator]}
\makeatother
\theoremheaderfont{\scshape}
\theorembodyfont{\rmfamily}
\theoremseparator{.}

\sectionfont{\normalfont\bfseries\normalsize}
\subsectionfont{\normalsize\normalfont\bfseries}

\makeatletter
\renewcommand\footnoterule{%
  \kern-3\p@
  \hrule\@width.4\columnwidth
  \kern2.6\p@}
  \makeatother

\DeclareUnicodeCharacter{2212}{-}
\begin{document}

\title{{\bf Alternative Fairness and Accuracy Optimization in Criminal Justice }%
\thanks{Shaolong Wu and Geshi Yeung are with Harvard University, and James Blume is with Massachusetts Institute of Technology. Contact: lorrywu@g.harvard.edu, jblume19@mit.edu, geshiyeung@g.harvard.edu. }}

\author{Shaolong Wu, James Blume, and Geshi Yeung}


\date{First Draft: December 2022, Revised: September, 2025\\[6mm]}

\maketitle

\vspace{-1.5cm}

\begin{abstract}
\onehalfspacing
\noindent
Algorithmic fairness has grown rapidly as a research area, yet key concepts remain unsettled, especially in criminal justice. We review group, individual, and process fairness and map the conditions under which they conflict. We then develop a simple modification to standard group fairness. Rather than exact parity across protected groups, we minimize a weighted error loss while keeping differences in false negative rates within a small tolerance. This makes solutions easier to find, can raise predictive accuracy, and surfaces the ethical choice of error costs. We situate this proposal within three classes of critique: biased and incomplete data, latent affirmative action, and the explosion of subgroup constraints. Finally, we offer a practical framework for deployment in public decision systems built on three pillars: need-based decisions, Transparency and accountability, and narrowly tailored definitions and solutions. Together, these elements link technical design to legitimacy and provide actionable guidance for agencies that use risk assessment and related tools.

\vspace{7mm}


\end{abstract}

\newpage
\epigraph{"You do not take a person who, for years, has been hobbled by chains and liberate him, bring him up to the starting line of a race and then say, "you are free to compete with all the others," and still justly believe that you have been completely fair."}{-President Lyndon B. Johnson}

\section{Introduction}
The use of algorithms has become increasingly pervasive in modern society, with many important decisions now being made by computers. As such, it is essential to ensure that these algorithms are designed and implemented in a fair and just manner. This paper will explore the concept of algorithmic fairness, considering the challenges and potential solutions to this important issue. By examining existing research, it will identify key implications for the future of computer science and consider how algorithmic fairness can be achieved.

Although the literature has taken great strides in both identifying instances of algorithmic unfairness and creating techniques to solve specific cases of unfairness, much work remains on defining ``fairness.'' This paper hopes to provide a general framework for the idea of fairness by offering guiding pillars that can be applied in a broad context. The paper shall proceed in the following sections:
\begin{enumerate}
    \item Popular definitions of fairness in machine learning
    \item Criticisms of the current framework
    \item Proposed Three Pillars of Fairness
\end{enumerate}

\newpage
\section{What is Algorithmic Fairness?}
Over the last decade, there has been a bewildering number of definitions of algorithmic fairness \cite{narayanan21fairness}. Making matters even more problematic, many of these proposed ideas of fairness are often incompatible with each other. Increasing fairness in one sense may decrease it in another. This paper shall focus on the three following dimensions:
\begin{enumerate}
    \item Group Fairness
    \item Individual Fairness
    \item Process Fairness
\end{enumerate}

\subsection{Group Fairness}
\subsubsection{Definitions}
Group fairness entails that an algorithm does not treat different demographic groups systematically differently. This idea of fairness was popularized after multiple real-world algorithms from criminal justice, corporate hiring, and credit ratings were found to systemically discriminate against minority candidates. While these algorithmic harms were often not malevolent in intent, their damage was tangible \cite{slaughter2020algorithms}. In particular, historically biased data often lead to a biased model. Algorithmic group fairness is often defined as having equal error rates between the desired groups. For example, a model for credit ratings should have an equal false positive rate between racial groups. In legal terms, group fairness is concerned with disparate impact.

There are also a few mathematical notions of group fairness that are based on probability \cite{soton467483}. First, there is demographic parity. This requires that there be the same proportion of individuals in any group receiving a positive outcome as the group's proportion of the population. Consider a binary classification setting, where a model has to make a prediction $\hat{Y} \in \{0, 1\}$, where $\hat{Y}=1$ means the model predicts an individual to have a positive outcome, while $Y=0$ means an individual is predicted to have a negative outcome. Then, let's say an individual has a group membership $S \in \{0,1\}$, where $S=0$ denotes membership to an underprivileged group while $S=1$ denotes membership to a privileged group. Then, group parity is achieved if $P(\hat{Y}=1|S=1)=P(\hat{Y}=1|S=0)$, and hence this can be measured by computing $\frac{P(\hat{Y}=1|S=1)}{P(\hat{Y}=1|S=0)}$. If the result is close to 1, there is group parity. However, if base rates are different, meaning that $P(Y=1|S=1)$ and $P(Y=1|S=0)$ are different, then even a classifier that never makes prediction errors (i.e. $\hat{Y}=Y$) will have a group parity measure that is not equal to 1, because $\frac{P(Y=1|S=1)}{P(Y=1|S=0)}$ is not equal to 1. 

Another mathematical definition of group fairness is called equalized odds, which requires the same false positive rate across groups as well as the same true positive rate across groups. This can be formulated as requiring $P(\hat{Y}=1|S=0,Y=y)$ and $P(\hat{Y}=1|S=1,Y=y)$ to be the same for $y \in \{0,1\}$. 

The third main definition of group fairness is an equal opportunity, in which only the true positive rate is required to be equal across groups. Formally, this means $P(\hat{Y}=1|S=0,Y=1)$ and $P(\hat{Y}=1|S=1,Y=1)$ need to be the same. Note this requirement is a subset of the requirement for equalized odds. 

The fourth definition of group fairness is calibration. It means that if the algorithm predicts the probability of a positive outcome to be $p$ for a set of individuals, then we should expect a $p$ portion of them to have a positive outcome. For example, for individuals predicted to have a high probability of recidivating, a large portion of them should actually have recidivated, which would mean that the algorithm is well-calibrated. Group fairness is achieved if calibration is held for different demographic groups (also called calibration within groups), meaning that for each demographic group, calibration should be held. Mathematically, in a binary classification setting, this means that $P(Y=1|S=0,\hat{Y}=1)$ and $P(Y=1|S=1,\hat{Y}=1)$ should be the same. 

These many definitions of group fairness mean that it is often up to the discretion of the algorithm designer to decide which fairness notion to adopt. Once a notion is adopted, the algorithm designer can then use many different methods to try to achieve group fairness. 

\subsubsection{Methods for achieving group fairness}
A naive way of achieving group fairness is through unawareness, which means that sensitive group attributes (such as race and gender) are to be excluded from the model. However, simply doing this can still easily result in suboptimal group fairness, since there may be features that correlate with the sensitive attributes and are still included in the model. If the model takes in those features to make predictions, and the sensitive attributes are historically tied to classification outcomes, individuals of different groups may still be treated differently. 

Another problem with unawareness is that group membership may sometimes offer valuable information, which the model would lose if group attributes are excluded from the model. \cite{dwork2012} discussed this utility by offering the example: suppose that in the culture of a protected class S, the most talented individuals would enter fields like science and engineering, while the less talented individuals enter fields like finance, and the trends are reversed in the general population. An organization hiring for talent that ignores group membership might select the subset of S most involved in economics and finance, but this is also the subset of S that is less talented. This is a poor outcome that arises from ignoring the group membership of S. Hence, ignoring sensitive attributes may not necessarily be good.

In addition to the naive method of unawareness, there are several other methods of pre-processing, which involve modifying the input data, that help achieve group fairness \cite{preprocessing}. First, unawareness may be replaced by suppression, where we remove not only the group membership attribute but also attributes that highly correlate with group membership, which would address the initial concern of unawareness. Second, the data may be pre-processed to remove bias against certain groups. For instance, the data may be ``massaged," where certain outcome labels are artificially changed. For example, one might turn a portion of the negative outcomes for an underprivileged group to become positive, which would cause the algorithm to more likely assign a positive label to an underprivileged group. The data may also be re-weighted, whereby individuals are assigned a weight and a larger weight could be assigned to an individual with a positive label from an underprivileged group during training. Last but not least, a class of algorithms called the disparate impact remover (DI remover) may be applied to the data, which can help achieve group fairness \cite{feldman}. Mathematically, the remover removes disparate impact: given a dataset $D=(X,X_n,Y)$ where $X$ is a protected attribute such as binary group membership, $X_n$ are the remaining attributes, and $Y$ is a binary outcome, the data set is considered to contain disparate impact if $\frac{P(Y=1|X=0)}{P(Y=1|X=1)} \leq \tau=0.8$, where $\tau$ is a tunable parameter depending on the need of the algorithm designer. This directly corresponds to the mathematical definition of demographic parity, where the closer $\tau$ is to 1 the stronger the demographic parity guarantee. A DI remover hence modifies the input data labels such that the input to the algorithm satisfies group parity. 

Besides pre-processing the data, the training procedure may also be modified, which is known as in-processing. A regularization term can be added to the loss function which penalizes treating members of different groups differently. For instance, during logistic regression, the loss function may be expressed as a weighted sum of binary cross-entropy loss, which reflects accuracy, and an unfairness function expressed as the difference between the average logistic regression score for the two groups \cite{RADOVANOVIC2022257}. The latter term helps control for how much the two groups can differ in terms of their regression score by the final model. 

Finally, the output of an algorithm could be post-processed to better abide by group fairness standards. For example, equalized odds post-processing is an algorithm that adds a simple post-processing step at the end to solve an optimization problem of achieving equal false positive and false negative rates, which involves flipping a certain amount of output labels \cite{NIPS2016_9d268236}. Another post-processing method is called reject option based classification (ROC), whereby the designer could set a threshold $\theta \in \{0.5,1\}$, predicting a positive label if confidence in that label exceeds $\theta$ and predicting a negative label if confidence is below $1-\theta$. If confidence is between $1-\theta$ and $\theta$, individuals from the underprivileged group are predicted a favorable outcome while individuals from the privileged group are predicted an unfavorable outcome \cite{preprocessing}. Note, however, that many of these examples are for a binary classification setting where individuals belong to one of two groups: privileged and underprivileged. Many of the in- and post-processing methods may not necessarily generalize to multi-class settings, and pre-processing may be a more general approach towards achieving group fairness.

\subsection{Individual Fairness}
Another intuitive definition of fairness comes from the notion of individual fairness. This can be characterized as whether similar individuals are treated similarly. This can be thought of as similar to the traditional racial color-blind arguments \cite{kleinberg2018algorithmic}. If a member of one group must satisfy a great threshold to achieve the same loan as another group, then the individual may perceive an injustice. Specifically, in a racial context, the individual may argue they are facing harm purely due to the color of their skin. This notion of fairness is the common assumption of many American anti-discrimination laws. In legal terms, individual unfairness can be thought of as disparate treatment. Mathematically, this may be formalized as the average difference between the label of an individual and the average label of the k-nearest neighbors of that individual based on non-sensitive attributes.    

\subsection{Process Fairness}
Unlike the other notions of fairness, which have been output-focused, process fairness concerns input fairness. In this framework, an algorithm gains legitimacy through having an open and transparent process \cite{grgic2016case}. While this idea is often under-discussed in computer science literature, it is very common in political science. At its core, fairness depends on whether people trust a given institution. An organization that is transparent both about its intentions and methods will be trusted more than an organization that obscures them. The value of process fairness is that it is robust to model errors and biased data because it does not depend on algorithmic outputs. 

\newpage


\newpage
\section{The Canonical Definition of Fairness and Its Critiques}

\subsection{The PAC setup for group fairness}

Similar to equal opportunity, a canonical setup for group fairness is to equalize false negative rates.

In the PAC setting, say the protected category variable is $V$, such that $V=\{v_1, \cdots, v_n\}$, where each of these is a particular realized value, such as Gender = $\{$male, female$\}$

The false negative rate is defined as:

\begin{equation}
    FN(h, v_i) :=  Pr_{(x,y) \sim P} [h(x) \neq -1 | x = v_i , y \neq +1 ]
\end{equation}

The goal is to ensure that the output hypothesis $h^{*}$ satisfies:

\begin{equation}
    FN(h, v_i) = FN(h, v_j), \forall i \neq j, 1 \leqslant i, j \leqslant n
\end{equation}

This is viewed as the constraint to the traditional loss minimization setting, where usually the goal is minimizing the sum of false negative rates and false positive rates in total: 

\[ \min \sum_{i=1}^{n} (\alpha FN(h, v_i) + \beta FP(h, v_i)) W_{i} \]

specifically, $W_{i}$ indicates the percent in the population whose protected trait turns out to be $v_i$, $\alpha$ is the loss assigned to false negative and $\beta$ is the loss assigned to false positive. 

Equalizing false negative rates can be viewed as the constraint to this loss minimization problem.

In specific settings, $\alpha$ and $\beta$ are tailored depending on how bad the two types of errors are. The relationship between the associated harm $\alpha$ and $\beta$ can be so balanced, such as

\[ \frac{\alpha}{\beta} >> 10  \] (1)

\[ \frac{\alpha}{\beta} << 0.1  \] (2)

For example, in the medical setting, doctors usually filter out those patients who likely have cancer and direct them to further rounds of tests. Thus, a false negative is a lot worse than a false positive, because not identifying latent cancer could delay therapies and intervention. In the case of a false positive, the doctor could simply not disclose the positive judgment directly and direct the patient to the next round of checking. This corresponds to (1).

However, in another setting, such as bank credit card applications. The influence of granting credits to individuals who potentially can't pay back does less social harm than withholding it from those who deserve it. Specifically, withholding credits from the lower social classes will have a significant negative impact on their household financial situations. This corresponds to (2).

In situation (2), the optimization problem is similar to:

\[ \min \sum_{i=1}^{n} \alpha FN(h, v_i) W_{i} \]

subject to equalized false negative rates. Previous literature, such as \cite{jakesch2022different} and \cite{grgic2016case}, have shown the need to flexibly address the balance between false negatives and positives. Being able to justify such a loss ratio $\frac{\alpha}{\beta}$ does indeed make the public audience trust the ethicality and process fairness of algorithms. 

In particular, in the setting of criminal justice, not prosecuting a suspect after a police investigation could pose nontrivial dangers to community safety, whereas excessively prosecuting all the potential suspects is a waste of judicial resources and may cause stigmatization or marginalization of communities. In such settings, striking the balance between $\alpha$ and $\beta$ seems inherently vague and tough.

Process fairness may require no arbitrary favoring of any group under the protected category in any step of the algorithm. This is a subtle definition. In some circumstances, it might not be feasible to achieve exactly. equalized false negative rates. One potential way to reconcile it is to give some tolerance bound, which is to make the false negative rejection rates across all the groups not differ beyond a bound. One practical situation could be to train a model that asks the predicted recidivism rate not to differ by 5 percent between all racial groups. This is a way to tackle the tradeoff between 'process fairness' and 'feasibility/accuracy'.

\subsection{Critiques to the Canonical Definition of Fairness}

Previous literature on group fairness has challenged its concept. The critiques are that the scheme is either insufficient or excessive.

\textbf{Critique 1.} \textit{Inherent biases in data}

The training data may contain unforeseen biases (e.g., in a complex field such as crime), and the algorithm may fail to correct them, as documented by \cite{mehrabi2021survey}. This phenomenon has been carefully examined empirically in \cite{chapman2022data} using a comprehensive data set from the official UK Crime API that provides data on policing's impact on crime rates.

Specifically, the underlying logic is similar to weight-adjusting mechanisms. More weights are assigned to the predictor that predicts there will be recidivism in the areas with historically high crime rates (especially if there's been a crime in the past several days, then the probability another crime will happen is considered very high by the algorithm without fairness adjustment). The learning theory behind "Near Repeat Theory" is that a crime incident triggers a temporary increase in crime rates nearby. Even with completely randomized synthesized "historical" data, the existing predictive policing algorithms lead to biased feedback loops that further confirm the assumed pattern. This is simply because more police in the areas with a prior crime makes crime in these areas more identifiable and further strengthens the need to police that area. 

Historically, in criminology \cite{gottfredson1987positive}, there's always a notion of positive criminology that believes crime to be the aggregate result of social issues such as psychological health problems, poverty, and social injustice. This thought is fundamentally against penalizing the "offenders" who are breaking laws as the last resort and not causing significant harm to the local community. From the positivist criminologist’s view, the unavoidable biases within the prior data set make the framework of fairness insufficient to address unfairness.

\bigbreak

\textbf{Critique 2.} \textit{Latent Affirmative Action}

Any group fairness definition ends up using affirmative action or similar logic, because one may inevitably end up giving an advantage to certain groups. As outlined by \cite{lagioia2022}, since different groups have different base rates, a system that has the same accuracy for different groups may fail to comply with group-parity standards. The authors analyzed the performance of Correctional Offender Management Profiling for Alternative Sanctions, also known as COMPAS, in evaluating defendants' risk profiles and classifying individuals as being at a high risk of recidivism if the system assigns the individual a score higher than a certain threshold based on the individual's criminal history, education, income level, family situation, etc. The authors found that by using the same threshold for different groups, individual fairness is achieved, in which the system would assign the same score to individuals with an equal likelihood of recidivism and therefore make the same classification for the two individuals. However, by setting different thresholds for the different groups, individuals with the same score may be classified differently, which violates individual fairness. Also, if the scores of different groups are calibrated, meaning that they are "equally correlated with the predicted classifications," then having different thresholds would lead to lowered accuracy for at least one of the groups. This accuracy-fairness tradeoff may be of concern to many. 

Overall, it appears that satisfying group fairness may result in unavoidable damage to individual fairness. Whether this is acceptable might depend on the quality of the input data: for example, if the input data is historically biased against certain groups, then having different thresholds allows the system to be calibrated such that predictions and probabilities can be aligned even with biased data. Also, depending on the policy goals, having different thresholds may be desired; ``the goal of increasing diversity or balancing access to education, types of jobs, or positions" may be some examples in which we might compromise individual fairness for certain policy goals [Lagioia]. Nonetheless, this paper illustrates how group fairness may be incompatible with individual fairness, and therefore one must be careful when applying group fairness metrics. As the authors argue, this requires ``discretionary value-based political choices" that statistical notions alone are insufficient to address.

Nonetheless, we know that individual fairness alone is not sufficient for ensuring a fair model either. As proposed by \cite{fleisher2021}, a model that classifies the same outcome for every individual would satisfy individual fairness (same treatment for similar individuals), but it is clearly unfair. Instead, \cite{speicher2018} has proposed an index for overall fairness that can be decomposed into two components: between-group and within-group unfairness. They acknowledged that improving components may be to the detriment of the other, but the decision of which component to prioritize might rest upon the user of the algorithm.     


\bigbreak
\textbf{Critique 3.} \textit{Vagueness of Fairness in Subgroups}

As raised in \cite{kearns2018preventing}, the intersection of demographic groups poses a challenge to achieving equalization of false negative rates across all protected categories. There are hundreds of different possible intersections of demographic variables, such as black homosexual women with a college education, and white heterosexual men with associate’s degrees. In most census or social surveys, there could be 6 different options under religion, 7 under race, 20 under country of origin, and 3 under sexual orientation. Naturally, this may introduce over one thousand categories with a considerable number of people in each of them. While the authors proposed an algorithm that relies on heuristics for learning to converge to the best subgroup-fair distribution over classifiers within polynomial time, it could be unclear how one should decide at what point we stop considering further intersection between subgroups. 

This subgrouping problem leads to the following three issues: 
\begin{enumerate}
    \item There is an insufficient moral justification for why it suffices to not consider the intersection of the protected features, but just consider group fairness alone.
    \item This practice is equivalent to introducing a thousand or even more linear constraints to a convex optimization problem, which could make the solution suboptimal or infeasible.
    \item To get a fair representation of the intersections between many different demographic subgroups, a large sample size may be needed for each intersection, which is difficult to obtain.
\end{enumerate}

\bigbreak
\textbf{Summary of Critiques}

While none of these critiques alone serve to make group fairness unworkable, put together, they raise concerns over whether to adopt certain notions of group fairness. Moreover, not every issue with group fairness may be solved with technical solutions alone, but may also require value-based decisions. For example, the choice of definition of group fairness will depend on the values of those affected by the algorithm. All this goes to show that any working idea of group fairness requires a value-based framework. The paper will explore this idea more fully in the last section.


\newpage
\section{Modified Alternative Definition of Fairness}
We develop this idea inspired by the related literature (\cite{ho2020affirmative} and \cite{schoeffer2022there}) ourselves as the following:

\[ \min \sum_{i=1}^{n} \alpha FN(h, v_i) W_{i} \]

subject to 

\[ |FN(h, v_i) - FN(h, v_j)| \leqslant \tau, \forall i \neq j, 1 \leqslant i, j \leqslant n \]

where $\alpha, \beta$ are the loss of false negatives and false positives, and $\tau$ is the tolerance bound for the difference in false negative rates.

This setup has two benefits: (i) by fine-tuning the parameter $\tau$, we can at least ensure that there's a feasible solution. (ii) Since the binding equalities are relaxed, the total accuracy (weighted sum of false negatives and false positives) can be higher.

The fundamental critique of this alternative setup is that designers of the algorithm may be well aware of the effect of which equations are the binding ones in the original setting. For example, in the bank credit application, we know that the probability of an African American person getting falsely denied is higher than a white American. Then by setting $\tau$ to be 5 percent, we are either making the African American applicant further disadvantaged. Or to the other extreme, if we decide to make the false negative rejection rate for African American persons to be 5 percent lower than other racial groups, we are essentially practicing 'affirmative action' based on race, which leads to significant legal controversy.

\newpage
\section{Individual fairness vs. group fairness}
As we have noted in previous sections, individual fairness and group fairness may sometimes be in conflict; doing better with the former might lead to unavoidable harm to the latter and vice versa. We will now examine the literature on this conflict in more detail. As argued by \cite{dwork2012}, premised on the classification setting where an algorithm needs to map an individual to a probability distribution over outcomes, individual fairness is achieved when the statistical distance between the distributions that individuals x and y are mapped is at most the distance between the two individuals, meaning that "the distributions over outcomes observed by x and y are indistinguishable up to their distance $d(x,y)$." Mathematically, this is known as the Lipschitz condition, where for a set of individuals $V$ and outcomes $A=\{0,1\}$, a mapping $M: V \rightarrow \Delta(A)$ satisfies the $(D,d)$-Lipschitz property if for every $x,y \in V$, $D(Mx, My) \leq d(x,y)$. D could be chosen as the statistical distance between two distributions P and Q, in which case $D(P, Q) = \frac{1}{2} \sum_{a\in A}|P(a)-Q(a)|$, while $d$ may denote the distance between two individuals on input attributes.

Achieving individual fairness can then be formulated as a linear program, where the expected loss of any arbitrary loss function $L: V \times A \rightarrow R$ is minimized subject to the constraint that the $(D,d)$-Lipschitz property is satisfied, meaning that the output distribution over outcomes for any two individuals differs by at most the distance of the two individuals. However, the researchers also proved that individual fairness (where the program is subject to the constraint of the Lipschitz property) implies group fairness if and only if the Wasserstein distance (a measure of the distance between two probability distributions) between the distributions of features between the two groups is small. This means that if the two groups share a similar distribution of features, individual fairness and group fairness can be achieved simultaneously. The opposite is also true: if the two groups share very different feature distributions, the two notions of fairness cannot be achieved simultaneously.

What happens then if the statistical distance between the two groups is large? How exactly does one balance individual fairness and group fairness? [Dwork, Hardt, Pitassi, Reingold, and Zemel] proposed an algorithm that attempts to balance the two by implementing \textit{fair affirmative action}. The algorithm relaxes the Lipschitz condition such that similar individuals from two different groups, $S$ and $T$ need not be treated similarly; the Lipschitz condition only needs to be held between individuals in group $S$ and between individuals in group $T$. This algorithm ensures demographic parity between S and T up to a bias $\epsilon$, meaning that $D(P, Q) = \frac{1}{2} \sum_{a\in A}|P(a)-Q(a)| \leq \epsilon$, where P and Q are the probability distributions over outcomes for the same individual with group membership S and T respectively. At the same time, individual fairness within a group is achieved, because the Lipschitz condition is satisfied for every pair $(x,y) \in (S\times S) \cup (T\times T)$. Hence, this work shows that by sacrificing individual fairness across groups, individual within groups and group fairness may be achieved simultaneously even if the feature distribution between different groups is very different.

To determine the empirical trade-off between individual and group fairness when the Wasserstein distance is large, \cite{soton467483} applied a disparate impact (DI) remover, which attempts to achieve group parity, on a real dataset \textit{Adult} \cite{asuncion2007uci}. The dataset consists of a binary sensitive attribute (sex or race), five non-sensitive attributes (e.g., age and education), and a binary outcome label of whether the individual's income exceeds 50K a year. The researchers found that a larger Wasserstein distance between the attribute distributions of the two groups (e.g. male vs. female) leads to a larger decrease in individual fairness after applying the DI remover. Individual fairness is also more likely to decrease if the large Wasserstein distance is due to a difference in mean rather than a difference in variance (as both can give the same Wasserstein distance). This confirms the intuition that it is difficult to achieve both group and individual fairness if the two groups are very different, especially if the mean rather than the variance of their attributes is different.

\newpage
\section{Viable Framework: Three Pillars of Fairness}
\label{sec:intro}
We have seen how individual and group fairness may be in conflict. In any order to reconcile these different ideas of fairness, it is crucial to have a guiding framework. Ideally, this framework provides general principles that can be applied to any organization and to address any problem of algorithmic unfairness.  In this paper, we propose the following Three Pillars as leading principles:
\begin{center}
\begin{enumerate}
    \item Need-based decisions
    \item Transparency and Accountability
    \item Narrowly Tailored Solutions and Definitions
\end{enumerate}
\end{center}
\subsection{Need-based decisions}
As outlined in \cite{srivastava2019mathematical}, while mathematical notions of fairness are important, it should be noted that fairness is inherently a value-based notion that may carry different meanings for different people or under different scenarios. For example, given data that may suffer from historical bias, setting different thresholds for classifications may help combat the bias in the data and help achieve group parity. However, if we are confident that all groups have a fair representation in the data, it may be best to set the same threshold across groups such that similar individuals, regardless of group membership, may be treated similarly. The decision as to which notion of fairness matters the most depends on the situation and the discretionary decision of the policymaker. There is thus not a one-size-fits-all notion of fairness that can be applied to all contexts, and we think future research in algorithmic fairness may continue to focus on how different specific scenarios may require different notions of fairness that best accommodate the needs of society in that area. 

\subsection{Transparency and Accountability}
Second, as we have explored in this review, there are a lot of varying definitions of fairness that may conflict with one another. Therefore, after a policy-maker chooses a notion of fairness and constructs a model that best satisfies the fairness requirement, it must be communicated clearly to the affected groups and society at large how the model was optimized for fairness. It should also be clarified how compromises have been made: for instance, how were potential conflicts between individual and group fairness balanced out by the system? This information should be communicated in an honest, transparent manner, and the use of mathematical notions should be used insofar as they convey essential information about the fairness criteria without unnecessarily rendering the interpretation of the model difficult for the layperson. This ensures that the public understands the nuances of how algorithmic fairness was achieved and could hold the system designer accountable for their decisions, which could have been masked if one simply labels their system as fair. 

At its core, algorithmic fairness is about correcting injustice. History has proven that the best way to advance this goal is through an open and transparent process. While some may argue that algorithms are too complicated and the topic of fairness too divisive for the process to be entirely open, it is, in fact, because of these reasons that transparency is important. The highly technical and politically controversial nature of algorithms is all the more reason that any solutions should face the marketplace of free ideas. Criticism over motivations and algorithmic adjustments serves to strengthen the process. This notion of process fairness is a leading motivator of the last pillar of "narrow definitions and solutions".

\subsection{Narrowly Tailored Solutions and Definitions}
The last and perhaps most important pillar that this paper proposes is that algorithmic fairness should rely on narrowly tailored definitions and solutions. Specifically, organizations should clearly define their definition of unfairness when designing their algorithm for a particular problem. The meaning of unfairness should be precisely communicated both mathematically and in a common language that everyday users can understand. This definition of fairness should also be accompanied by historical justifications for algorithmic adjustments to this specific problem. While this is a high threshold, narrowly tailored definitions of fairness for a specific problem are the best practice to maintain trust and legitimacy in the process. In addition, narrow solutions serve a similar role. Any algorithmic solutions should be justified for the specific problem that it is designed to solve. This limitation serves not only to make the problem more workable but also to enhance process fairness. 

\subsubsection{Why Tailored Definitions are Needed}
To address any specific case of algorithmic unfairness, it is crucial to have a tailored definition of unfairness. An organization using a too broad or ill-conceived definition will struggle not only to create a technical solution to the problem but may face what political scientists refer to as ``mission creep.'' Mission creep is when there is a gradual shift away from the initial mission. In this case, the fear is that algorithm creators may be able to commit arbitrary changes by justifying them with some notion of fairness. In this way, broad definitions of algorithmic unfairness may lead to a decrease in algorithmic accountability and transparency.

	Likewise, there is no good reason to assume that every problem will require the same definition of fairness. For example, if the creator of an algorithm is a corporation, it may have different goals of fairness than if the creator were a benign social planner \cite{jakesch2022different}. It would be both unreasonable and likely undesirable to require every corporation to play the role of the ultimate arbiter of fairness. In this way, any definition of fairness should be limited by the incentives and power of the algorithmic creator. More broadly, a large assortment of different problems may all fall under the same umbrella term of unfairness and it would make little sense to use the same definition in every context.

	Finally, a tailored definition of unfairness makes technical solutions more feasible. The main limitation of technical solutions to algorithmic unfairness has been requiring a model to do too much. Different notions of fairness carry with them their trade-offs. There is almost always no solution that can satisfy every definition of fair. In light of this, a context-specific, tailored definition of fairness also has the benefit of being the most workable. 

\subsubsection{Why Tailored Solutions are Needed}
Even once a narrowly tailored definition is agreed upon for a given problem, it is equally important to have a narrowly tailored solution. Cookie-cutter solutions risk legal challenges from US anti-discriminatory laws \cite{slaughter2020algorithms}. In general, the US requires that any racial discrimination be narrowly tailored and that any race-neutral options to have been exhausted. In this sense, any attempt to create racial algorithmic fairness must be specific to the context of the problem. Put more concretely, in the context of college applications, each algorithm must be based on that school’s specific history. A school in the South that historically discriminated against black students until the 1970s should have a different model than that of a historically black college.

	The University of Chicago Law Review paper, ``Affirmative Algorithms: The Legal Grounds for Fairness as Awareness'' \cite{ho2020affirmative} argues that narrowly tailored solutions not only are the most legally viable method of implementing non-race blind algorithms but are also the best at achieving their goals. The more a model is adjusted to its unique history and training data flaws, the more likely it is to accomplish true fairness. The paper goes on to argue that each instance of non-racially blind algorithms should be thoroughly justified by social science investigations of the discriminatory past. Likewise, a larger history of discrimination may vindicate a larger readjustment of the model.

	The need for a tailored solution goes beyond legal positioning. From the perspective of output-based legitimacy, a tailored solution will yield better outcomes. The problems of algorithmic harm are in no way uniform, so it would make little sense to expect their solutions to be uniform. Over-applying the same method in every scenario risks miscalibration in the other direction. Similarly, from an input-based legitimacy like process fairness, being more conservative and careful with applying model readjustments serves to maintain public trust. To most of the public, the methods of machine learning and algorithms are black boxes. Any tinkering on sensitive issues like race will likely be viewed with great suspicion. To these ends, assuring the public that any such modifications will be narrowly tailored around the unique history of the problem will serve to ease public mistrust.

 \section{Conclusion}
As algorithmic fairness moves from an academic idea to real-world applications, a general framework must be established. This framework should provide universal principles that organizations can follow to ensure fairness in their own algorithms. While there will always exist trade-offs between different systems, a proper framework can act as a guiding star to navigate historical discrimination and injustice.  This paper hopes to lay the foundations of such a framework with our Three Pillars Model.

Future extensions of our framework could provide more mathematically rigorous analysis of the trade-offs between different types of unfairness. Likewise, more research can be done to integrate the political science idea of ``legitimacy'' into our model. Other extensions of this paper may include examples of practical applications of the Three Pillars Model. Specifically, future papers may show proposed analysis of real-world problems using the Three Pillars Model in comparison to other frameworks. We hope this could offer a holistic framework to tackle the complex issues of algorithmic unfairness.

\newpage
\section{Appendix}
\label{sec:intro}


\printbibliography

\end{document}